\documentclass{article}

\usepackage[preprint]{neurips_2026}

\usepackage[utf8]{inputenc} % allow utf-8 input
\usepackage[T1]{fontenc}    % use 8-bit T1 fonts
\usepackage{hyperref}       % hyperlinks
\usepackage{url}            % simple URL typesetting
\usepackage{booktabs}       % professional-quality tables
\usepackage{amsfonts}       % blackboard math symbols
\usepackage{nicefrac}       % compact symbols for 1/2, etc.
\usepackage{microtype}      % microtypography
\usepackage{xcolor}         % colors
\usepackage{graphicx}
\usepackage{wrapfig} 
\usepackage{cleveref}
\usepackage[table]{xcolor} 
\usepackage{bm}
\usepackage{multirow}

\usepackage{enumitem}
\usepackage{parskip}

\title{Choosing the Right Representations: Task-Aware Fusion for Multimodal Materials Language Models}

\author{%
\hspace{30pt} Huu Binh Ta \\
\hspace{30pt} University of Virginia \\
\hspace{30pt} \texttt{nzj6jt@virginia.edu}
\And
\hspace{15pt} Alex Morehead \\
\hspace{15pt} Lawrence Berkeley National Laboratory \\
\hspace{15pt} \texttt{acmwhb@lbl.gov}
\AND 
\hspace{-25pt} N. Benjamin Erichson \\
\hspace{-25pt} Lawrence Berkeley National Laboratory \\
\hspace{-25pt} \texttt{erichson@lbl.gov}
\And 
\hspace{-19pt} Ferdinando Fioretto \\
\hspace{-19pt} University of Virginia \\
\hspace{-19pt} \texttt{fioretto@virginia.edu} 
\AND 
Sandeep Madireddy \\
Argonne National Laboratory \\
\texttt{smadireddy@anl.gov} \\
}

\begin{document}

\maketitle

\begin{abstract}
Multimodal language models for materials science rely on effective mechanisms for incorporating three-dimensional structural representations. MatterChat addresses this challenge using a large bridge module between a pretrained structural encoder and the LLM, together with a separate contrastive alignment stage.
We instead ask whether this bridge can adaptively select and combine representations from multiple structural encoders according to the target property.
We find that supervised and generative material encoders provide complementary information for property prediction. We found that, for Zatom-1, this information is not concentrated in the final layer. Instead, the most informative representations occur at different depths depending on the target property.
This motivates Task-Conditioned Attention Fusion (TCAF), a lightweight attention-based bridge that uses task-conditioned latent queries to combine CHGNet features with intermediate Zatom-1 representations. The bridge is trained directly on downstream prediction tasks, allowing representation selection and material--language alignment to be learned jointly without separate contrastive pretraining. Across classification and regression tasks, TCAF outperforms MatterChat, with up to 47.8\% lower RMSE, while using substantially fewer trainable parameters and halving the overall training time. Further experiments incorporating UMA-S demonstrate TCAF’s flexibility across encoder combinations, with the three-encoder configuration achieving a highest macro-average classification accuracy of 89.61\%.
\end{abstract}

\section{Introduction}
In silico materials discovery has traditionally relied on high-fidelity computational methods such as density functional theory \cite{kohn1996density} and ab initio molecular dynamics \cite{marx2000ab}. While accurate, these methods are computationally expensive and difficult to scale across large chemical spaces \cite{tang2025matterchat}. Graph-based machine-learning interatomic potentials (MLIPs) \cite{batzner20223,gasteiger2021gemnet,batatia2025foundation} address this limitation by learning representations of local atomic environments \citep{cretu2026zatom} and have substantially accelerated material-property prediction. However, these models operate on atomic structures alone and cannot directly incorporate natural-language queries or scientific context \cite{birhane2023science}.

Large language models (LLMs) offer a complementary way to interact with materials data through natural language \cite{kim2024large,cavanagh2026smileyllama}. Most existing approaches represent materials using one-dimensional text formats, including chemical formulas \cite{kim2024large}, SMILES strings \cite{cavanagh2026smileyllama,weininger1988smiles,ta2026search}, textual descriptions \cite{ock2023catalyst}, and crystallographic information files \cite{antunes2024crystal}. These representations can encode structural information, but processing them as token sequences does not provide the geometric inductive biases used by models designed explicitly for three-dimensional atomic environments. As a result, text-based models generally remain less accurate than dedicated graph models for quantitative property prediction \cite{ock2023catalyst}.This motivates multimodal models that connect 3D material structures with LLMs.

MatterChat \cite{tang2025matterchat} takes an important step in this direction by combining material graphs with natural-language queries. It uses CHGNet \cite{deng2023chgnet} as a pretrained structural encoder and connects its atom-level representations to an LLM through a BLIP-2-style bridge \cite{li2023blip}. The bridge consists of alternating self- and cross-attention layers and is trained in two stages: contrastive pretraining first aligns the structural and textual modalities, followed by supervised fine-tuning for downstream property prediction. While effective, this design introduces a large number of trainable parameters and requires a separate alignment stage. It also relies on a single structural encoder, although different encoders may provide complementary information for different material properties.

This raises a more general question: \emph{rather than committing to a single structural representation, can the bridge adaptively select and combine representations according to the target property?} We investigate this question using two structurally different encoders: CHGNet, a supervised MLIP, and Zatom-1 \cite{morehead2026zatom}, a general-purpose multimodal flow-matching model \cite{miller2024flowmm} for three-dimensional molecules and materials. Our analysis shows that the two encoders provide complementary information for property prediction. We also find that, for Zatom-1, useful information is not concentrated in the final layer. Different intermediate layers are most informative for different material properties, suggesting that a fixed representation from the generative encoder is unnecessarily restrictive.

These observations motivate Task-Conditioned Attention Fusion (TCAF). TCAF uses a lightweight attention-based bridge model based on Perceiver \cite{jaegle2021perceiver, hawthorne2022general} architecture to combine representations from CHGNet with intermediate representations from Zatom-1. The bridge operates through cross-attention \cite{vaswani2017attention}, with latent queries conditioned on the target property using a mechanism inspired by prompt tuning \cite{lester2021power}. These task-conditioned queries allow the bridge to change which encoder and which Zatom-1 layers it attends to for different material-property queries. Thus, instead of aligning each structural representation with language in advance, the bridge learns representation selection, fusion, and material-language alignment directly from the downstream training objective.
Interestingly, this formulation also removes the need for a separate contrastive pretraining stage. The resulting model is trained in a single stage while using fewer trainable parameters than MatterChat. Across classification and regression tasks, TCAF improves predictive performance over MatterChat and reduces the overall training time by half. We further evaluate TCAF with encoder combinations incorporating UMA-S \cite{wood2026family}, demonstrating its flexibility in integrating additional structural representations.

The paper presents three key contributions:
\begin{enumerate}[leftmargin=*, nosep]

% \item \textbf{Task-conditioned representation fusion.} We introduce TCAF, a lightweight Perceiver-based bridge that dynamically combines supervised and generative structural representations according to the target material property. TCAF learns representation selection and material--language alignment directly from the downstream objective, eliminating the need for contrastive pretraining.

\item \textbf{Task-conditioned representation fusion.} We introduce TCAF, a single-stage framework with a lighter bridge model than the heavy two-stage MatterChat. TCAF additionally replaces static structural alignment with task-conditioned latent queries, allowing the model to dynamically route attention across supervised features, leading to a consistent improvement on all tasks.

\item \textbf{Analysis of generative structural representations.} We show that property-relevant information in Zatom-1 is distributed across depth and that the most informative layers vary by material property. We further show that these representations complement those learned by CHGNet.

\item \textbf{Empirical performance and efficiency.} TCAF achieves the highest classification accuracy across all evaluated tasks and the lowest RMSE on all three regression tasks, reducing regression RMSE by up to 47.8\% while using at least 80\% fewer trainable parameters and half the training time.

\end{enumerate}

\section{Method: Task-conditioned Representation Fusion (TCAF)}

\begin{figure*}[!t]
    \centering
    \includegraphics[width=0.85\textwidth]{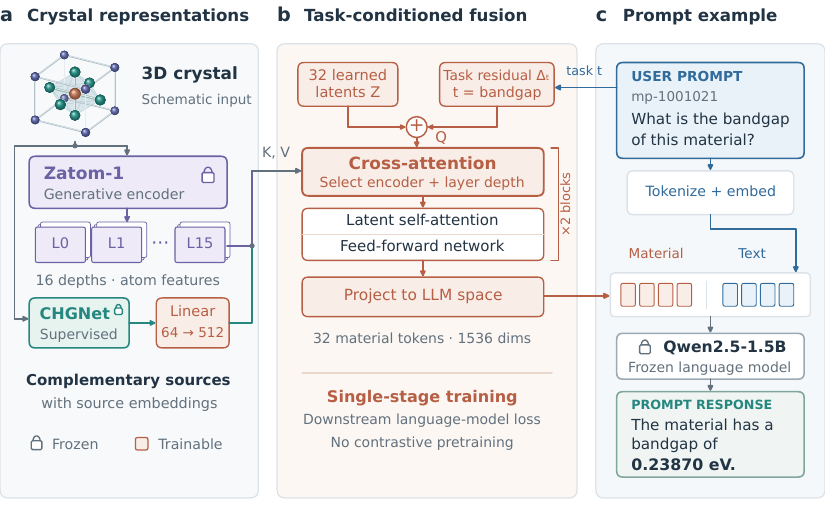}
    \vspace{-0.2cm}
    \caption{Overview of our proposed Task-Conditioned Attention Fusion (TCAF) methodology. \textbf{(a)} Frozen domain encoders (Zatom-1 and CHGNet) extract complementary atom-level representations from a crystal structure. Here, features from all 16 Zatom-1 layers are retained, and CHGNet features are projected from 64 to 512 dimensions. \textbf{(b)} The property requested by the prompt selects a learned task residual, which is added to 32 latent queries. Two Perceiver blocks apply cross-attention to the structural features, followed by latent self-attention and a feed-forward network. These task-conditioned queries allow the bridge to weight encoder sources and layer depths differently for each property. \textbf{(c)} The fused latents are projected into the language-model embedding space and combined with prompt embeddings. A frozen language model then generates the response. The trainable components are optimized jointly using a downstream language-modeling objective.}
    % \label{fig:tcaf_overview}
    \label{fig:method}
\end{figure*}

% This section presents a multimodal framework for material property prediction that integrates 3D atomic structures with LLMs through a single-stage training pipeline. By leveraging a lightweight attention-based bridge and a novel task-conditioned fusion mechanism, our approach completely eliminates the need for contrastive pretraining while effectively combining complementary structural representations.

%\textbf{Architecture.}
The proposed framework combines 3D crystal structures with natural-language queries for material-property prediction, as illustrated in \Cref{fig:method}. The input structure is processed by two frozen structural encoders: Zatom-1, a generative model, and CHGNet, a supervised machine-learning interatomic potential. CHGNet produces 64-dimensional atom embeddings, which we project to 512 dimensions to match the Zatom-1 representations.
Representations from both encoders are passed to a trainable Perceiver bridge model, which combines them through cross-attention. The resulting latent tokens are appended to the natural-language prompt and passed to a frozen LLM. 
%
%We use Qwen2.5-1.5B \cite{hui2024qwen2} throughout this work.

\textbf{Single-Stage Training.}
MatterChat uses a parameter-heavy BLIP-2-style bridge and a separate contrastive pretraining stage to align structural and textual representations. We instead use a lightweight attention-based bridge that compresses the variable-length sequence of atom embeddings into 32 latent tokens. Its cross-attention is mask-aware, ensuring that padded atoms do not contribute to the resampling operation.
The bridge is trained directly through the language-modeling objective of the frozen LLM on the downstream prediction tasks. This allows material-language alignment to be learned together with the task objective and removes the need for separate contrastive pretraining. As a result, the complete model is trained in a single stage, reducing the overall training time by half.

\begin{table}[t]
    \centering
    \begin{minipage}{0.39\textwidth}
        \centering

        \footnotesize 
        \caption{Top-3 best-performing Zatom-1 layers for each classification task, identified via linear probing.}
        \setlength{\tabcolsep}{1pt}
        \begin{tabular*}{\textwidth}{@{\extracolsep{\fill}} l c c c @{}}
        \toprule
        \textbf{Task} & \textbf{1st Best} & \textbf{2nd Best} & \textbf{3rd Best}\\
        \midrule
        Metallic & L0 & L1 & L2 \\
        Direct Bandgap & L2 & L7 & L10 \\
        Stability & L0 & L2 & L6 \\
        Exp Ob & L1 & L2 & L0 \\
        Is Magnetic & L0 & L1 & L2 \\
        Magnetic Order & L0 & L2 & L1 \\
        % \rowcolor{gray!20}
        % SearchDiff & $391.8 \pm 6.1$ & $0.66 \pm 0.01$ & 2 \\
        % \midrule
        % CDD        & 22              & 0.62 \\
        % % \rowcolor{gray!20}
        % SearchDiff & $448.6 \pm 9.0$ & $0.70 \pm 0.00$ \\
        % \midrule
        % CDD        & 33              & 0.58 \\
        % % \rowcolor{gray!20}
        % SearchDiff & $484.8 \pm 6.3$ & $0.75 \pm 0.00$ \\
        \bottomrule
        \end{tabular*}
        
        % \captionof{table}{Performance under increasing SA thresholds (comparison with CDD) on the QM9 dataset. CDD results reported from \cite{cardei2025constrained}.}
        \label{tab:linear_probing_top_3}
    \end{minipage}
    \hfill
    \begin{minipage}{0.58\textwidth}
        \centering

        \footnotesize 
        \caption{Classification accuracy comparison between individual Zatom-1 layers and the supervised CHGNet baseline. Bold values indicate the best performance for each task.}
        \setlength{\tabcolsep}{1pt} % minimum col space
        \begin{tabular*}{\textwidth}{@{\extracolsep{\fill}} l c c c c @{}}
        \toprule
        \textbf{Task} & \textbf{L0} & \textbf{L2} & \textbf{L15} & \textbf{CHGNet}\\
        \midrule
        Metallic {\scriptsize[$\uparrow$]} & 0.7758 & \textbf{0.8331} & 0.8218 & \textit{0.6986} \\
        Direct Bandgap {\scriptsize[$\uparrow$]} & \textit{0.8424} & \textbf{0.8751} & 0.8708 & 0.8698 \\
        Stability {\scriptsize[$\uparrow$]} & 0.8274 & 0.8322 & \textit{0.7620} & \textbf{0.8695} \\
        Exp Ob {\scriptsize[$\uparrow$]} & 0.8064 & \textbf{0.8171} & \textit{0.7796} & 0.8112 \\
        Is Magnetic {\scriptsize[$\uparrow$]} & 0.8727 & 0.8596 & \textit{0.8194} & \textbf{0.9218} \\
        Magnetic Order {\scriptsize[$\uparrow$]} & 0.8144 & 0.8170 & \textit{0.8004} & \textbf{0.8466} \\
        \bottomrule
        \end{tabular*}
        
        \label{tab:Layer_performance}
    \end{minipage}
\end{table}

\textbf{Task-Conditioned Attention Fusion.}
Our analysis shows that property-relevant information in Zatom-1 is distributed across network depth rather than concentrated in its final layer (see \Cref{tab:linear_probing_top_3}). Therefore, we aim to design a method that can pass the representations from multiple Zatom-1 layers to the Perceiver together with the projected CHGNet representation. These representations form the keys and values of the cross-attention operation, while 32 learned latent tokens act as queries. Each Perceiver block applies cross-attention followed by latent self-attention and a feed-forward network \cite{vaswani2017attention}.
To make this fusion depend on the target property, we "dynamically modify" the latent queries using a mechanism inspired by prompt tuning \cite{lester2021power}. For each material property, we learn a query residual that is added to the initial latent tokens before the attention blocks. The property specified by the natural-language prompt determines which residual is used. The resulting task-conditioned queries allow the bridge to adapt its attention to both encoder source and Zatom-1 depth.
%We call the resulting framework Task-Conditioned Attention Fusion (TCAF). By combining complementary structural representations according to the target property, TCAF performs representation selection, fusion, and material--language alignment within the same lightweight bridge.
While we focus on Zatom-1 and CHGNet in the primary configuration, TCAF is not restricted to this encoder pair. We additionally evaluate encoder combinations incorporating UMA-S to assess the flexibility of the fusion framework, as reported in \Cref{sec:experiment}.

\section{Experimental Analysis}
\label{sec:experiment}

We evaluate our framework on the materials science instruction-tuning dataset introduced with MatterChat \cite{tang2025matterchat}. We use a frozen Qwen2.5-1.5B language model across all experiments. We compare TCAF with MatterChat and a Projector baseline, which uses an MLP-based bridge to map structural representations into 32 latent tokens without attention.

\textbf{Identifying Property-Discriminative Layers.}
We first investigate which layers of the frozen Zatom-1 encoder contain the most property-relevant information. Because the final layers are optimized for its pretraining objective, intermediate representations may be better suited for downstream property prediction. We test this by training a logistic regression probe on mean-pooled representations from each of the 16 Zatom-1 layers.
% \begin{wrapfigure}[19]{r}{0.6\columnwidth}
%     \centering
%     \includegraphics[width=0.58\columnwidth]{figures/classification_accuracy_lineplot.pdf}
%     \caption{Classification accuracy of TCAF versus other methods.}
%     \label{fig:classification_accuracy}
% \end{wrapfigure}
As shown in \Cref{tab:linear_probing_top_3}, property-relevant information is concentrated in the earlier layers of Zatom-1. Layer 0 (L0) performs best for most classification tasks, while Layer 2 (L2) ranks among the top three layers across all evaluated properties. We next evaluate L0 and L2 in the full LLM pipeline and compare them with the final Zatom-1 layer (L15) and the supervised CHGNet baseline. As shown in \Cref{tab:Layer_performance}, L15 consistently performs worst, indicating that the final layer of the generative encoder might not have the most useful representation for material-property prediction. L2, in contrast, is competitive with CHGNet across several tasks. However, no single Zatom-1 layer consistently outperforms CHGNet across all properties. These results show that the two encoders provide complementary representations and motivate their adaptive fusion.

\begin{wrapfigure}[15]{r}{0.51\columnwidth}
    \centering
    \vspace{-0.3cm}
    \includegraphics[width=0.5\columnwidth]{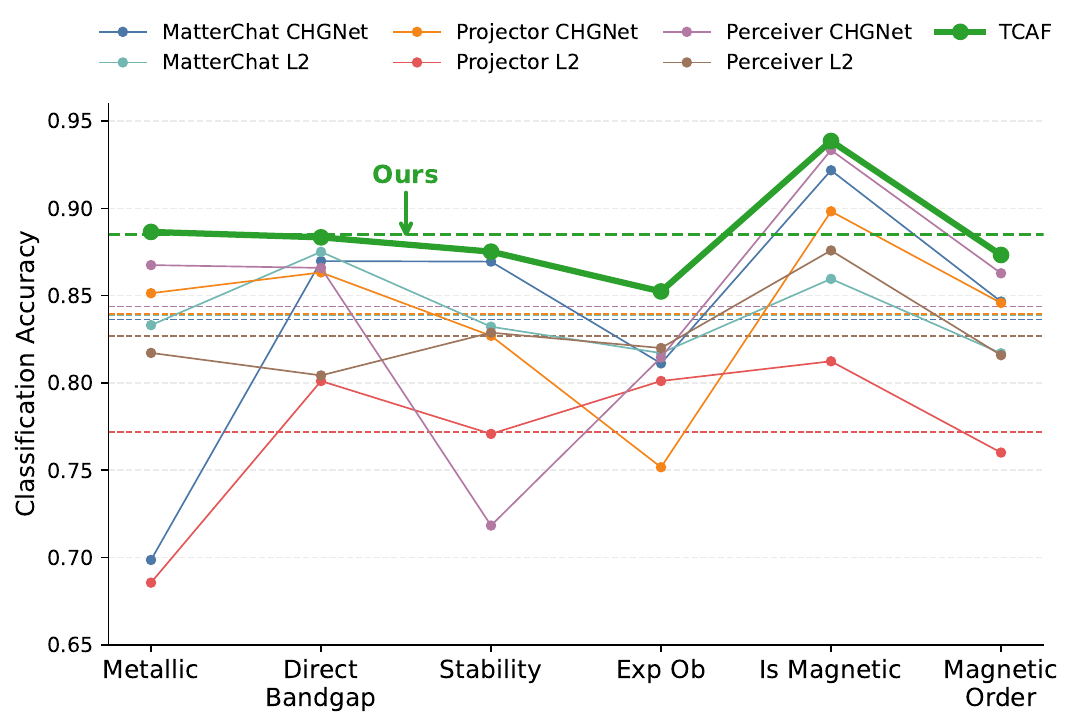}
    \vspace{-0.3cm}
    \caption{Comparision of classification accuracy. %TCAF (ours) outperforms baselines. 
    Dashed lines indicate average accuracy over tasks.}
    \label{fig:classification_accuracy}
\end{wrapfigure}

\textbf{Task-Conditioned Attention Fusion Performance.} 
\Cref{fig:classification_accuracy} compares the classification accuracy of seven methods across all tasks. TCAF achieves the highest accuracy on each task, showing the benefit of task-conditioned fusion. In contrast, the Projector L2 baseline performs worst overall, suggesting that a simple MLP is insufficient to connect the generative representations with the LLM.
Perceiver CHGNet achieves the second-highest average accuracy and performs similarly to MatterChat across most tasks. Unlike MatterChat, however, the Perceiver bridge is trained in a single stage and uses substantially fewer trainable parameters (20M vs. more than 100M of MatterChat).
\Cref{tab:regression_tcaf} reports regression performance together with the number of invalid generations, where an output is considered invalid if the LLM fails to produce a parseable floating-point value. TCAF achieves the lowest RMSE on all three regression tasks (up to 47.8\% lower than MatterChat in "Energy Above Hull") while producing only a small number of invalid outputs. This suggests that task-conditioned fusion improves both predictive accuracy and the stability of the generated outputs. Additional experimental details are provided in Appendix A and B, and fusion and task-conditioning ablations, together with cross-attention analyses, are provided in Appendix C.

\begin{table}[t]
\centering
\footnotesize
\caption{Regression performance of TCAF versus other methods.}
\setlength{\tabcolsep}{6pt}
\renewcommand{\arraystretch}{1.15}
\begin{tabular}{@{} l c c c c c c @{}}
\toprule
% \textbf{Approach} & \textbf{Metallic} & \textbf{Direct Bandgap} & \textbf{Stability} & \textbf{Exp Ob} & \textbf{Is Magnetic} & \textbf{Magnetic Order} & \textbf{Formation Energy} & \textbf{} & \textbf{Energy Above Hull} & \textbf{} & \textbf{Bandgap} & \textbf{} \\
% \textbf{Method} & \textbf{Metallic {\scriptsize[$\uparrow$]}} & \textbf{Direct Bandgap {\scriptsize[$\uparrow$]}} & \textbf{Stability {\scriptsize[$\uparrow$]}} & \textbf{Exp Ob {\scriptsize[$\uparrow$]}} & \textbf{Is Magnetic {\scriptsize[$\uparrow$]}} & \textbf{Magnetic Order {\scriptsize[$\uparrow$]}} \\
% \midrule
% Matterchat CHGNet & 0.6986 & 0.8698 & 0.8695 & 0.8112 & 0.9218 & 0.8466 \\
% Matterchat L2  & 0.8331 & 0.8751  & 0.8322  & 0.8171  & 0.8596  & 0.8170  \\
% \midrule
% Projector CHGNet  & 0.8514 & 0.8633 & 0.8270 & 0.7517 & 0.8983 & 0.8458 \\
% Projector L2  & 0.6856 & 0.8011 & 0.7708 & 0.8011 & 0.8124 & 0.7601 \\
% \midrule
% Perceiver CHGNet  & 0.8675 & 0.8659 & 0.7183 & 0.8146 & 0.9334 & 0.8628 \\
% Perceiver L2  & 0.8172 & 0.8043 & 0.8288 & 0.8200 & 0.8759 & 0.8159 \\
% \rowcolor{gray!20}
% TCAF  & 0.8865 & 0.8834 & 0.8752 & 0.8524 & 0.9386 & 0.8733 \\
% \toprule
\multirow{2}{*}{\textbf{Method}} 
& \multicolumn{2}{c}{\textbf{Formation Energy}}
& \multicolumn{2}{c}{\textbf{Energy Above Hull}}
& \multicolumn{2}{c}{\textbf{Bandgap}} \\
\cmidrule(lr){2-3}\cmidrule(lr){4-5}\cmidrule(lr){6-7}
& \textbf{No. Invalid {\scriptsize[$\downarrow$]}} & \textbf{RMSE {\scriptsize[$\downarrow$]}}
& \textbf{No. Invalid {\scriptsize[$\downarrow$]}} & \textbf{RMSE {\scriptsize[$\downarrow$]}}
& \textbf{No. Invalid {\scriptsize[$\downarrow$]}} & \textbf{RMSE {\scriptsize[$\downarrow$]}} \\
\midrule
MatterChat CHGNet & 2 & 0.2944 & 380 & 0.2560 & 1 & 0.6731 \\
MatterChat L2  & 258 & 1.5861 & 13 & 0.3061 & 571 & 1.1968 \\
\midrule
Projector CHGNet  & 62 & 0.2226 & 79 & 0.2795 & 2 & 0.7681 \\
Projector L2  & 594 & 0.6702 & 894 & 0.4713 & 1742 & 1.6955  \\
\midrule
Perceiver CHGNet  & 55 & 0.2095 & 486 & 0.2363 & 46 & 0.9921 \\
Perceiver L2  & 1039 & 0.4896 & 143 & 0.3277 & 94 & 1.2794 \\
\rowcolor{orange!20}
TCAF  & 1 & \textbf{0.1775} & 28 & \textbf{0.1336} & 1 & \textbf{0.5356} \\
\bottomrule
\end{tabular}
\label{tab:regression_tcaf}
\end{table}

\textbf{Generalization Across Structural Encoders.} To assess the generalizability and modularity of TCAF when incorporating new representations, we integrate UMA-S alongside existing encoders. As summarized in \Cref{tab:uma_s_classification_performance}, applying TCAF across various encoder combinations yields consistent performance gains over baseline strategies. Both \textit{Matterchat UMA-S} and \textit{Perceiver UMA-S}) achieve lower macro scores of 86.44\% and 85.84\%. Combining two encoders improves performance: \textit{TCAF CHGNet+UMA-S} and \textit{TCAF Zatom+UMA-S} reach macro scores of 88.82\% and 89.02\%, with the latter achieving the highest accuracy on \textit{Direct gap} (88.61\%) and \textit{Magnetic} (95.51\%).
Combining all three encoders (\textit{TCAF Zatom+CHGNet+UMA-S}) achieves the best overall result, with a top macro score of \textbf{89.61\%}. It secures the highest performance in four out of six tasks. These results show that TCAF can effectively integrate multiple distinct representations to boost overall accuracy.

\begin{table}[t]
\centering
\footnotesize
\caption{Performance comparison of UMA-S encoder on classification tasks.}
\setlength{\tabcolsep}{4.5pt}
\renewcommand{\arraystretch}{1.15}
\begin{tabular}{@{} l c c c c c c c @{}}
\toprule
\textbf{Algorithm} & \textbf{Metal} & \textbf{Direct gap} & \textbf{Stable} & \textbf{Observed} & \textbf{Magnetic} & \textbf{Mag. order} & \textbf{Macro} \\
 & {\scriptsize[$\uparrow$]} & {\scriptsize[$\uparrow$]} & {\scriptsize[$\uparrow$]} & {\scriptsize[$\uparrow$]} & {\scriptsize[$\uparrow$]} & {\scriptsize[$\uparrow$]} & {\scriptsize[$\uparrow$]} \\
\midrule
Matterchat UMA-S & 87.43 & 85.84 & 86.71 & 79.10 & 92.71 & 86.86 & 86.44 \\
Perceiver UMA-S & 88.53 & 87.35 & 79.02 & 81.22 & 91.83 & 87.09 & 85.84 \\
\midrule
TCAF CHGNet+UMA-S & 90.81 & 87.50 & 88.14 & 83.16 & 95.17 & 88.14 & 88.82 \\
TCAF Zatom+UMA-S & 86.80 & \textbf{88.61} & 88.57 & 86.04 & \textbf{95.51} & 88.61 & 89.02 \\
\rowcolor{orange!20}
TCAF Zatom+CHGNet+UMA-S & \textbf{90.97} & 87.99 & \textbf{88.83} & \textbf{86.73} & 94.49 & \textbf{88.64} & \textbf{89.61} \\
\bottomrule
\end{tabular}
\label{tab:uma_s_classification_performance}
\end{table}

\section{Conclusion}

We introduced Task-Conditioned Attention Fusion (TCAF), a lightweight framework for multimodal material-property prediction that removes the need for a separate contrastive pretraining stage. Using task-conditioned latent queries within a Perceiver bridge, TCAF dynamically combines intermediate representations from the generative model Zatom-1 with representations from the supervised interatomic potential CHGNet. Our analysis shows that property-relevant information in Zatom-1 is distributed across network depth and varies with the target property, supporting the use of adaptive rather than fixed structural representations. Across classification and regression tasks, TCAF improves predictive performance while using substantially fewer trainable parameters. These results suggest that multimodal material--language models can benefit from selecting and combining complementary structural representations directly during downstream training.

\begin{ack}
This research used resources of the National Energy Research Scientific Computing Center, a DOE Office of Science User Facility supported by the Office of Science of the U.S. Department of Energy (DOE) under Contract No. DE-AC02-05CH11231, using the AI4Sci@NERSC award NERSC DDRERCAP0036206 awarded to AM. NBE would also like to acknowledge support from the U.S. Department of Energy, Office of Science, Office of Advanced Scientific Computing Research, through the RAPIDS Institute under the Scientific Discovery through Advanced Computing (SciDAC) program, under Contract No. DE-AC02-05CH11231 at Lawrence Berkeley National Laboratory. 
%The views, opinions, and/or findings expressed are those of the authors and should not be interpreted as representing the official views or policies of the U.S. Government.
%
This work was also partially supported by NSF grants CAREER-2401285, RI-2533631, RI-2334936, the LaCross Institute Fellowship and DARPA under Contract No.~\#HR0011252E005. 
This material is based upon work supported by the National Science Foundation Graduate Research Fellowship Program under Grant No 2234693. 
%Any opinions, findings, and conclusions or recommendations expressed in this material are those of the authors and do not necessarily reflect the views of the National Science Foundation. 
The authors also acknowledge the Research Computing at the University of Virginia. 
% Any opinions, findings, and conclusions or recommendations expressed in this material are those of the authors and do not necessarily reflect the views of the NSF or DARPA.
The views, opinions, and/or findings expressed are those of the authors and should not be interpreted as representing the official views or policies of the U.S. Government.
\end{ack}

\clearpage
\bibliographystyle{unsrt}
\bibliography{references}

\begin{thebibliography}{10}

\bibitem{kohn1996density}
Walter Kohn, Axel~D Becke, and Robert~G Parr.
\newblock Density functional theory of electronic structure.
\newblock {\em The journal of physical chemistry}, 100(31):12974--12980, 1996.

\bibitem{marx2000ab}
Dominik Marx and Jurg Hutter.
\newblock Ab initio molecular dynamics: Theory and implementation.
\newblock {\em Modern methods and algorithms of quantum chemistry}, 1(301-449):141, 2000.

\bibitem{tang2025matterchat}
Yingheng Tang, Wenbin Xu, Jie Cao, Weilu Gao, Steven Farrell, Benjamin Erichson, Michael~W Mahoney, Andy Nonaka, and Zhi~Jackie Yao.
\newblock A multimodal large language model for materials science.
\newblock {\em Nature Machine Intelligence}, pages 1--14, 2026.

\bibitem{batzner20223}
Simon Batzner, Albert Musaelian, Lixin Sun, Mario Geiger, Jonathan~P Mailoa, Mordechai Kornbluth, Nicola Molinari, Tess~E Smidt, and Boris Kozinsky.
\newblock E (3)-equivariant graph neural networks for data-efficient and accurate interatomic potentials.
\newblock {\em Nature communications}, 13(1):2453, 2022.

\bibitem{gasteiger2021gemnet}
Johannes Gasteiger, Florian Becker, and Stephan G{\"u}nnemann.
\newblock Gemnet: Universal directional graph neural networks for molecules.
\newblock {\em Advances in neural information processing systems}, 34:6790--6802, 2021.

\bibitem{batatia2025foundation}
Ilyes Batatia, Philipp Benner, Yuan Chiang, Alin~M Elena, D{\'a}vid~P Kov{\'a}cs, Janosh Riebesell, Xavier~R Advincula, Mark Asta, Matthew Avaylon, William~J Baldwin, et~al.
\newblock A foundation model for atomistic materials chemistry.
\newblock {\em The Journal of chemical physics}, 163(18), 2025.

\bibitem{cretu2026zatom}
Miruna Cretu, Alex Abrudan, Antonia Panescu, Tynan Perez, Rishabh Anand, N~Benjamin Erichson, Michael~W Mahoney, Samuel Blau, Joseph Jacobson, Rafael G{\'o}mez-Bombarelli, et~al.
\newblock Zatom-2: Multitask pretraining on atomistic data for generative modeling across domains.
\newblock {\em arXiv preprint arXiv:2610.11454}, 2026.

\bibitem{birhane2023science}
Abeba Birhane, Atoosa Kasirzadeh, David Leslie, and Sandra Wachter.
\newblock Science in the age of large language models.
\newblock {\em Nature Reviews Physics}, 5(5):277--280, 2023.

\bibitem{kim2024large}
Seongmin Kim, Yousung Jung, and Joshua Schrier.
\newblock Large language models for inorganic synthesis predictions.
\newblock {\em Journal of the American Chemical Society}, 146(29):19654--19659, 2024.

\bibitem{cavanagh2026smileyllama}
Joseph~M Cavanagh, Kunyang Sun, Andrew Gritsevskiy, Dorian Bagni, Yingze Wang, Thomas~D Bannister, and Teresa Head-Gordon.
\newblock Smileyllama: Modifying large language models for directed chemical space exploration.
\newblock {\em Nature Computational Science}, pages 1--14, 2026.

\bibitem{weininger1988smiles}
David Weininger.
\newblock Smiles, a chemical language and information system. 1. introduction to methodology and encoding rules.
\newblock {\em Journal of chemical information and computer sciences}, 28(1):31--36, 1988.

\bibitem{ta2026search}
Huu~Binh Ta, Michael Cardei, Alvaro Velasquez, and Ferdinando Fioretto.
\newblock Search-augmented masked diffusion models for constrained generation.
\newblock {\em arXiv preprint arXiv:2602.02727}, 2026.

\bibitem{ock2023catalyst}
Janghoon Ock, Chakradhar Guntuboina, and Amir Barati~Farimani.
\newblock Catalyst energy prediction with catberta: unveiling feature exploration strategies through large language models.
\newblock {\em ACS Catalysis}, 13(24):16032--16044, 2023.

\bibitem{antunes2024crystal}
Luis~M Antunes, Keith~T Butler, and Ricardo Grau-Crespo.
\newblock Crystal structure generation with autoregressive large language modeling.
\newblock {\em Nature Communications}, 15(1):10570, 2024.

\bibitem{deng2023chgnet}
Bowen Deng, Peichen Zhong, KyuJung Jun, Janosh Riebesell, Kevin Han, Christopher~J Bartel, and Gerbrand Ceder.
\newblock Chgnet as a pretrained universal neural network potential for charge-informed atomistic modelling.
\newblock {\em Nature Machine Intelligence}, 5(9):1031--1041, 2023.

\bibitem{li2023blip}
Junnan Li, Dongxu Li, Silvio Savarese, and Steven Hoi.
\newblock Blip-2: Bootstrapping language-image pre-training with frozen image encoders and large language models.
\newblock In {\em International conference on machine learning}, pages 19730--19742. PmLR, 2023.

\bibitem{morehead2026zatom}
Alex Morehead, Miruna Cretu, Antonia Panescu, Rishabh Anand, Maurice Weiler, Tynan Perez, Samuel Blau, Steven Farrell, Wahid Bhimji, Anubhav Jain, et~al.
\newblock Zatom-1: Towards a multimodal foundation model for 3d molecules and materials.
\newblock {\em arXiv preprint arXiv:2602.22251}, 2026.

\bibitem{miller2024flowmm}
Benjamin~Kurt Miller, Ricky~TQ Chen, Anuroop Sriram, and Brandon~M Wood.
\newblock Flowmm: Generating materials with riemannian flow matching.
\newblock {\em arXiv preprint arXiv:2406.04713}, 2024.

\bibitem{jaegle2021perceiver}
Andrew Jaegle, Sebastian Borgeaud, Jean-Baptiste Alayrac, Carl Doersch, Catalin Ionescu, David Ding, Skanda Koppula, Daniel Zoran, Andrew Brock, Evan Shelhamer, et~al.
\newblock Perceiver io: A general architecture for structured inputs \& outputs.
\newblock {\em arXiv preprint arXiv:2107.14795}, 2021.

\bibitem{hawthorne2022general}
Curtis Hawthorne, Andrew Jaegle, C{\u{a}}t{\u{a}}lina Cangea, Sebastian Borgeaud, Charlie Nash, Mateusz Malinowski, Sander Dieleman, Oriol Vinyals, Matthew Botvinick, Ian Simon, et~al.
\newblock General-purpose, long-context autoregressive modeling with perceiver ar.
\newblock In {\em International Conference on Machine Learning}, pages 8535--8558. PMLR, 2022.

\bibitem{vaswani2017attention}
Ashish Vaswani, Noam Shazeer, Niki Parmar, Jakob Uszkoreit, Llion Jones, Aidan~N Gomez, {\L}ukasz Kaiser, and Illia Polosukhin.
\newblock Attention is all you need.
\newblock {\em Advances in neural information processing systems}, 30, 2017.

\bibitem{lester2021power}
Brian Lester, Rami Al-Rfou, and Noah Constant.
\newblock The power of scale for parameter-efficient prompt tuning.
\newblock In {\em Proceedings of the 2021 conference on empirical methods in natural language processing}, pages 3045--3059, 2021.

\bibitem{wood2026family}
Brandon Wood, Misko Dzamba, Xiang Fu, Meng Gao, Muhammed Shuaibi, Luis Barroso-Luque, Kareem Abdelmaqsoud, Vahe Gharakhanyan, John Kitchin, Daniel Levine, et~al.
\newblock Uma: A family of universal models for atoms.
\newblock {\em Advances in Neural Information Processing Systems}, 38:129391--129427, 2026.

\bibitem{hu2021unit}
Ronghang Hu and Amanpreet Singh.
\newblock Unit: Multimodal multitask learning with a unified transformer.
\newblock In {\em 2021 IEEE/CVF International Conference on Computer Vision (ICCV)}, pages 1419--1429. IEEE, 2021.

\bibitem{vielzeuf2018centralnet}
Valentin Vielzeuf, Alexis Lechervy, St{\'e}phane Pateux, and Fr{\'e}d{\'e}ric Jurie.
\newblock Centralnet: a multilayer approach for multimodal fusion.
\newblock In {\em European Conference on Computer Vision}, pages 575--589. Springer, 2018.

\end{thebibliography}

\appendix
\onecolumn

\section{Experimental Details}

\subsection{Dataset and Evaluation Protocol}
Our experiments utilized the instruction-tuning dataset introduced by the MatterChat \cite{tang2025matterchat} framework. The dataset from Matterchat \footnote{https://zenodo.org/records/18735961} encompasses nine distinct material property prediction tasks. Following data cleaning and post-processing, the corpus yields a total of 1,285,144 structure-instruction pairs. The exact distribution of training and test samples across the nine tasks is detailed in Table \ref{tab:dataset_distribution}. 
To address the issue of generative degeneration where language models output unparseable text or have decimal point missing results, we report the number of invalid samples and RSME for the remaining samples.

\begin{table}[h]
\centering
\footnotesize
\caption{Dataset distribution across classification and regression tasks.}
\begin{tabular}{@{} l r r r @{}}
\toprule
\textbf{Task} & \textbf{Train} & \textbf{Test} & \textbf{Total} \\
\midrule
Is Metal & 128,609 & 14,290 & 142,899 \\
Direct Bandgap & 128,609 & 14,290 & 142,899 \\
Stability & 128,609 & 14,290 & 142,899 \\
Experimentally Observed & 128,609 & 14,290 & 142,899 \\
Is Magnetic & 127,821 & 14,204 & 142,025 \\
Magnetic Order & 128,537 & 14,289 & 142,826 \\
Bandgap & 128,609 & 14,290 & 142,899 \\
Energy Above Hull & 128,609 & 14,290 & 142,899 \\
Formation Energy & 128,609 & 14,290 & 142,899 \\
\midrule
\textbf{Total} & \textbf{1,156,621} & \textbf{128,523} & \textbf{1,285,144} \\
\bottomrule
\end{tabular}
\label{tab:dataset_distribution}
\end{table}

\subsection{Architecture Specifications}
The crystal encoders and language model are frozen throughout all
experiments. We use Qwen2.5-1.5B-Instruct with a hidden dimension of 1,536.
The structural encoders are configured as follows:
\begin{itemize}[leftmargin=*]
    \item \textbf{CHGNet:} A supervised interatomic model producing
    64-dimensional per-atom embeddings.
    \item \textbf{Zatom-1:} A 16-layer generative transformer producing
    512-dimensional per-atom embeddings at each layer.
    \item \textbf{UMA-S:} A universal interatomic model producing
    128-dimensional per-atom embeddings.
\end{itemize}

We compare three bridge architectures that map structural features
into 32 tokens in the language-model embedding space:
\begin{itemize}[leftmargin=*]
    \item \textbf{Q-Former (MatterChat baseline):} Initialized from
    BERT-base with 12 layers, 12 attention heads, a hidden dimension
    of 768, a feed-forward width of 3,072, and a dropout rate of 0.1.
    It uses 32 learned query tokens and applies cross-attention
    to the structural features every two layers.

    \item \textbf{MLP Projector:} Uses mask-aware adaptive average
    pooling to obtain 32 structural tokens, followed by a linear
    projection to 1,536 dimensions, a GELU activation, a second
    linear projection, and LayerNorm.

    \item \textbf{Perceiver (TCAF):} Uses 32 learned latents of
    dimension 768, two resampler blocks, 8 attention heads,
    a feed-forward width of 3,072, and a dropout rate of 0.1.
    A final projection maps the latents to 1,536 dimensions.
\end{itemize}

\begin{figure*}[!t]
    \centering
    \includegraphics[width=0.85\textwidth]{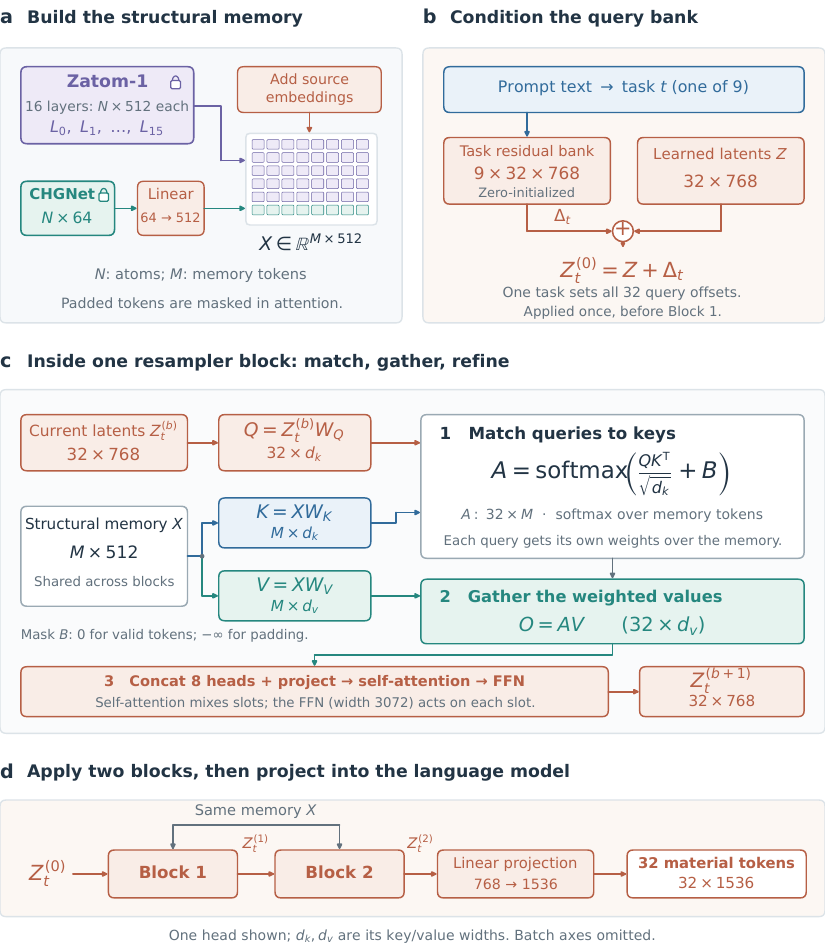}
    \vspace{-0.2cm}
    \caption{Detailed view of the TCAF bridge. \textbf{(a)} Features from the frozen structural encoders are mapped to a common width of 512 and combined with learned source embeddings to form the structural memory $X$.
    \textbf{(b)} The prompt determines the task index $t$, which selects one
    of nine learned residual matrices. The selected residual is added to
    the shared latent queries, giving $Z_t^{(0)} = Z + \Delta_t$.  This conditioning occurs once, before the first resampler block.  \textbf{(c)} Each block forms queries from the current latents and keys and values from $X$. Query--key similarities determine attention weights over the memory, with padded positions excluded before softmax. The weighted sum $O = AV$ retrieves structural information for each query. Outputs from eight heads are concatenated and projected, followed by self-attention across the 32 latent slots and a feed-forward network applied to each slot. \textbf{(d)} Two blocks update the latents while reusing the same structural memory. A final projection produces 32 material tokens of width 1536 for the language model. Batch dimensions are omitted, and the attention equations show one head. Here, $M$ is the memory length, and $d_k$ and $d_v$ are the per-head key and value widths.}
    \label{fig:tcaf_bridge_details}
\end{figure*}

Figure~\ref{fig:tcaf_bridge_details} details the computation within TCAF.
In our primary Zatom-1 + CHGNet configuration, CHGNet features are
projected from 64 to 512 dimensions and combined with representations
from all 16 Zatom-1 layers to form the structural memory $X$.
Learned 512-dimensional source embeddings distinguish the encoder
sources and layer depths. These embeddings are initialized from
$\mathcal{N}(0, 0.02)$.

Task conditioning uses a zero-initialized residual bank of shape
$9 \times 32 \times 768$, with one residual matrix per prediction task.
The task index $t$ is inferred exclusively from the prompt text;
dataset metadata is not used for task selection. The selected
residual is added to the shared learned latents before the first block:
\[
    Z_t^{(0)} = Z + \Delta_t,
    \qquad
    Z, \Delta_t \in \mathbb{R}^{32 \times 768}.
\]
Each task therefore adjusts the full bank of 32 queries. Zero
initialization makes the initial queries identical across tasks,
while training learns property-specific adjustments.

Within each resampler block, the current latents provide the queries,
and the structural memory provides the keys and values.
Mask-aware cross-attention retrieves structural information while
excluding padded positions. Latent self-attention and a feed-forward
network then update the retrieved representations. Both blocks use
the same memory $X$, but the second block operates on the updated
latents from the first. This lets the requested property influence
which encoder features and layer depths the bridge emphasizes.
The final 32 material tokens are combined with prompt embeddings
and passed to the frozen language model.

\subsection{Training Hyperparameters}
To ensure computational efficiency, all structural embeddings were precomputed and cached on local temporary file systems prior to training. Training was conducted in BF16 mixed precision across 12 distributed hardware accelerator ranks. 

Models were trained using the AdamW optimizer with $\beta = (0.9, 0.999)$ and a weight decay of 0.05. We applied a maximum gradient clipping norm of 1.0. The learning rate followed a linear warm-up schedule from $10^{-6}$ to a peak of $10^{-4}$ over the first 500 optimization steps, subsequently decaying via a cosine schedule down to $10^{-5}$. With a local batch size of 32 and a gradient accumulation step of 4 across 12 ranks, the effective global batch size was maintained at 1,536. Both pretraining (where applicable) and fine-tuning phases were executed for 7,500 steps. During inference, the maximum generation length was strictly restricted to 256 tokens.

\subsection{Trainable Parameters}

To further highlight the computational advantages of our proposed framework, Table \ref{tab:trainable_params} details the number of trainable parameters required by each bridging architecture. The foundational MatterChat framework imposes a substantial computational burden, as its Q-Former bridge necessitates a two-stage training process. During the contrastive pretraining phase (Stage 1), MatterChat requires approximately 197 million to 201 million trainable parameters. Subsequently, during the supervised fine-tuning phase (Stage 2), it still maintains roughly 118 million to 122 million trainable parameters.

In contrast, the Projector, Perceiver, and our proposed TCAF architectures bypass the resource-intensive pretraining phase, operating exclusively within a single-stage pipeline. While the Projector baseline utilizes the fewest parameters (under 4 million), it severely underperforms in regression and generative stability, as shown in previous sections. TCAF, built upon the Perceiver architecture, requires only 20,772,224 trainable parameters. This effectively demonstrates that TCAF is exceptionally parameter-efficient: it utilizes nearly an order of magnitude fewer trainable parameters compared to MatterChat's Stage 1 and reduces the Stage 2 parameter count by roughly $80\%$, all while completely eliminating the multi-stage training paradigm and achieving better predictive performance.

\begin{table}[h]
\centering
\footnotesize
\caption{Comparison of trainable parameters across different bridge architectures. For single-stage models (Projector, Perceiver, and TCAF), the training is conducted end-to-end without a separate pretraining phase.}
\setlength{\tabcolsep}{12pt}
\renewcommand{\arraystretch}{1.15}
\begin{tabular}{@{} l c c @{}}
\toprule
\textbf{Method} & \textbf{Stage 1} & \textbf{Stage 2} \\
\midrule
MatterChat CHGNet & 197,756,477 & 118,020,666 \\
MatterChat L2 & 201,885,245 & 122,149,434 \\
\midrule
Projector CHGNet & \multicolumn{2}{c}{2,463,744} \\
Projector L2 & \multicolumn{2}{c}{3,151,872} \\
Perceiver CHGNet & \multicolumn{2}{c}{20,163,840} \\
Perceiver L2 & \multicolumn{2}{c}{20,507,904} \\
\rowcolor{gray!20}
TCAF & \multicolumn{2}{c}{20,772,224} \\
\bottomrule
\end{tabular}
\label{tab:trainable_params}
\end{table}

\subsection{Hardware Infrastructure}
All computational workloads, including feature extraction, training, and evaluation, were executed on a large-scale high-performance computing (HPC) cluster. Specifically, the distributed environment utilized a single compute node equipped with 12 advanced hardware accelerator ranks (architecturally equivalent to the Intel Data Center GPU Max Series). Each accelerator rank features 64 GB of High Bandwidth Memory (HBM), providing substantial memory capacity to efficiently handle large-scale multimodal training. Furthermore, these accelerators offer native hardware support for \texttt{bfloat16} precision, which we leveraged to maximize training throughput and optimize memory bandwidth without the need for gradient scaling.

\section{Data Input-Output and Caching}

During distributed training, processing high-resolution structural data often introduces severe I/O bottlenecks. Initially, storing precomputed embeddings as individual files per structure, resulting in over 142,000 tiny files totaling approximately 68 GB. This caused massive read latencies and metadata contention on the shared distributed file system. At a global batch size of 1536, the concurrent read requests from multiple ranks overwhelmed the file system, leading to highly erratic and stalled step times.

To reduce this extreme metadata contention, we implemented an optimized caching pipeline: All per-layer and supervised embeddings were first pre-packed into a single, large contiguous archive. Furthermore, to mitigate I/O bottlenecks and metadata contention typical in distributed file systems, all precomputed embeddings were staged directly to node-local temporary file systems prior to training execution. This node-local staging effectively decoupled the training loop from the shared cluster filesystem, allowing all per-step reads to access high-speed RAM-backed storage. This caching strategy stabilized step times and accelerated overall training throughput by over an order of magnitude, rendering the training process entirely independent of shared file system congestion.

\section{Additional Experiment}

% \textbf{Approach} & \textbf{Metallic} & \textbf{Direct Bandgap} & \textbf{Stability} & \textbf{Exp Ob} & \textbf{Is Magnetic} & \textbf{Magnetic Order} & \textbf{Formation Energy} & \textbf{} & \textbf{Energy Above Hull} & \textbf{} & \textbf{Bandgap} & \textbf{} \\
% \textbf{Method} & \textbf{Metallic {\scriptsize[$\uparrow$]}} & \textbf{Direct Bandgap {\scriptsize[$\uparrow$]}} & \textbf{Stability {\scriptsize[$\uparrow$]}} & \textbf{Exp Ob {\scriptsize[$\uparrow$]}} & \textbf{Is Magnetic {\scriptsize[$\uparrow$]}} & \textbf{Magnetic Order {\scriptsize[$\uparrow$]}} \\
% \midrule
% Matterchat & 0.6986 & 0.8698 & 0.8695 & 0.8112 & 0.9218 & 0.8466 \\
% Weighted Sum & 0.8766 & 0.8605 & 0.8736 & 0.8454 & 0.9212 & 0.8634 \\
% Concatenation & 0.8653 & 0.8715 & 0.8567 & 0.8493 & 0.9134 & 0.8555 \\
% TCAF no task conditioning  & 0.8832 & 0.8787 & 0.8701 & 0.8369 & 0.9013 & 0.8712 \\
% TCAF  & 0.8865 & 0.8834 & 0.8752 & 0.8524 & 0.9386 & 0.8733 \\
% \toprule

To further evaluate the efficacy of the proposed TCAF architecture, we experiment with several fusion variants and ablation settings. Specifically, we implement \textit{Concatenation} (inspired by \cite{hu2021unit}) and \textit{Weighted Sum} (inspired by \cite{vielzeuf2018centralnet}) as alternative mechanisms to combine the Zatom and CHGNet embeddings while retaining the task-conditioning module. Additionally, we evaluate an ablated version of our model, \textit{TCAF (no task cond)}, to isolate the impact of task-aware latent queries on overall performance.

\begin{figure*}[t]
    \centering
    \includegraphics[width=1\textwidth]{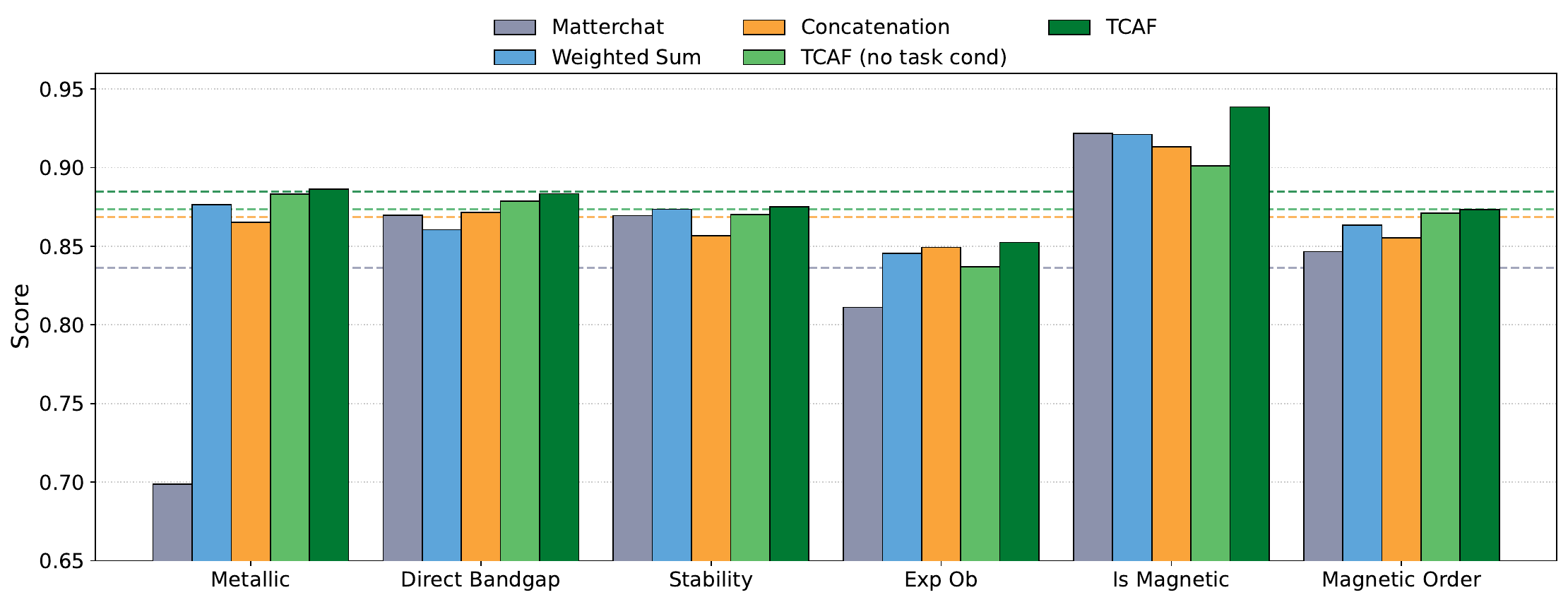}
    \caption{Performance comparison of multimodal fusion strategies across six classification tasks. The grouped bars evaluate the baseline Matterchat, basic fusion mechanisms (Weighted Sum, Concatenation), and the proposed TCAF architecture with and without task conditioning. Dashed horizontal lines indicate the macro-average score of each respective method across all tasks, highlighting the overall performance gains achieved by the fully conditioned TCAF.}
    \label{fig:performance_grouped_bar}
\end{figure*}

As illustrated in \Cref{fig:performance_grouped_bar}, the fully conditioned TCAF consistently achieves the best performance across all evaluated material property prediction tasks. Conversely, the baseline Matterchat model exhibits the weakest performance across the board, despite utilizing more learnable parameters and requiring longer training times. Among the alternative embedding combination strategies, naive concatenation yields the poorest results. Interestingly, the macro-average performance of the Weighted Sum variant and the unconditioned TCAF are nearly identical (0.8735 and 0.8736, respectively). This plateau suggests that while basic structural fusion techniques can aggregate multi-scale features effectively, dynamic task-specific conditioning is essential to achieve state-of-the-art performance.

To understand the mechanism behind TCAF's performance improvements, we visualize the internal cross-attention weights of the Perceiver blocks to analyze how the model aggregates multimodal features.

\begin{figure*}[t]
    \centering
    \includegraphics[width=1\textwidth]{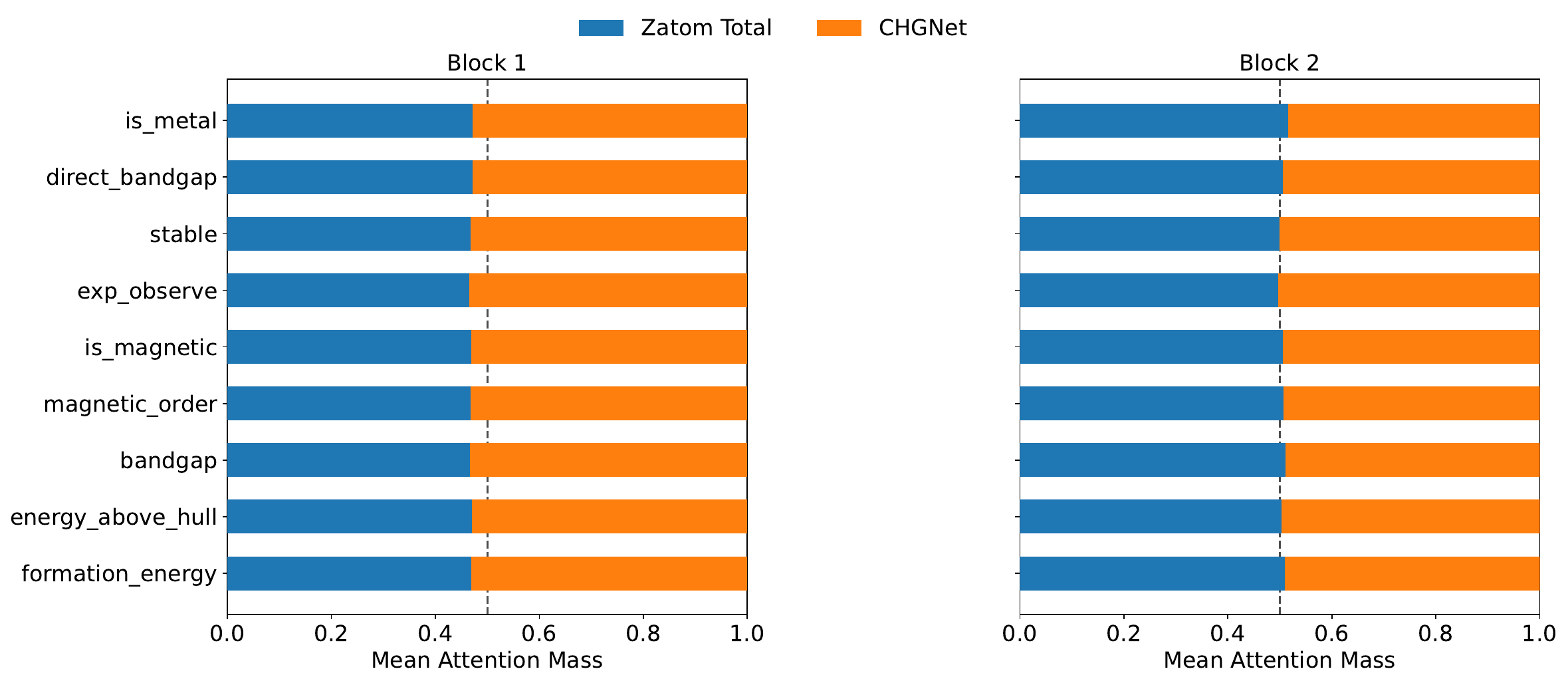}
    \caption{Macro-level encoder contribution (Zatom vs. CHGNet) for the baseline Perceiver model without task conditioning. The stacked bars illustrate the mean cross-attention mass allocated to the aggregated 16 Zatom layers compared to the single CHGNet representation across Block 1 and Block 2.}
    \label{fig:notaskcond_stacked_bar}
\end{figure*}

\begin{figure*}[t]
    \centering
    \includegraphics[width=1\textwidth]{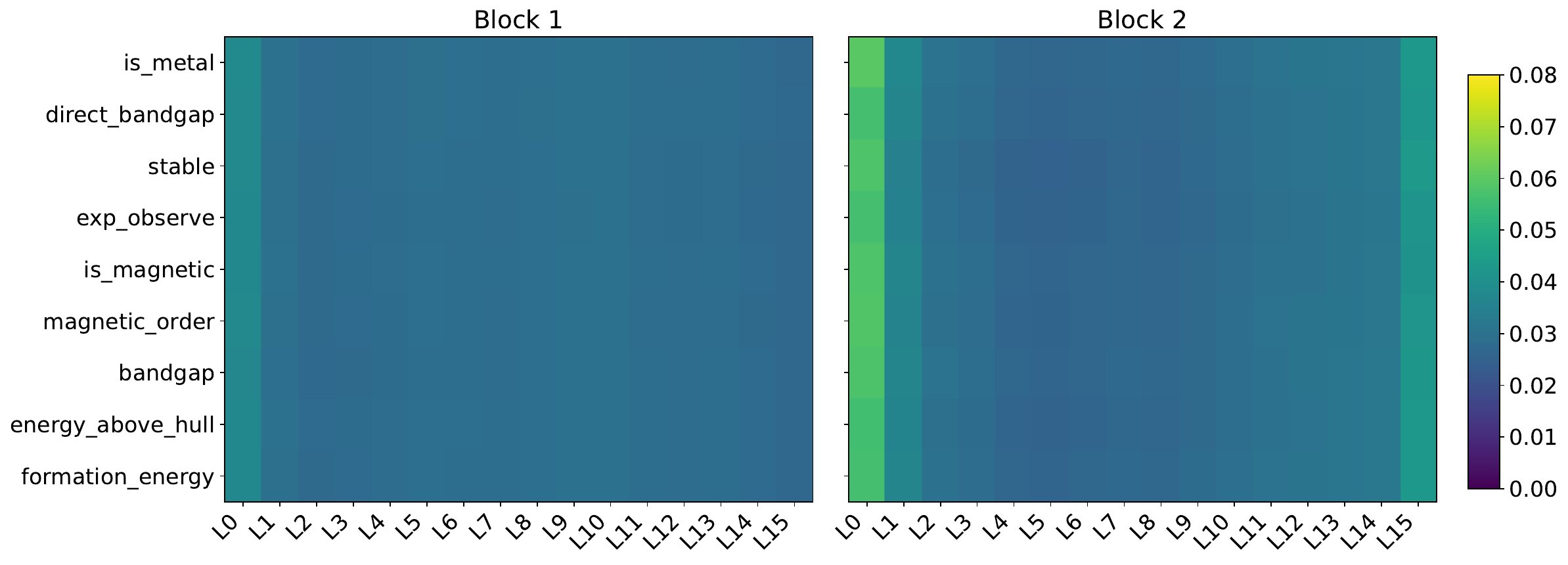}
    \caption{Micro-level cross-attention distribution across the 16 Zatom intermediate layers for the baseline Perceiver model without task conditioning. Color intensity indicates the mean attention mass assigned to each depth layer for different property prediction tasks.}
    \label{fig:notaskcond_zatom_heatmap}
\end{figure*}

In the unconditioned baseline, the absence of task-specific latent queries results in a nearly completely static attention distribution. As demonstrated in \Cref{fig:notaskcond_stacked_bar}, the macro-level allocation of attention mass between the aggregated Zatom layers and the CHGNet embedding remains rigid and identical across all nine target properties. This lack of adaptability extends to the micro-level routing; the heatmap in \Cref{fig:notaskcond_zatom_heatmap} reveals that the unconditioned model uniformly extracts features from the exact same Zatom depth layers regardless of the physical task being evaluated. This behavior indicates an attention collapse, where the model defaults to a generalized, average representation rather than specializing its feature extraction.

\begin{figure*}[t]
    \centering
    \includegraphics[width=1\textwidth]{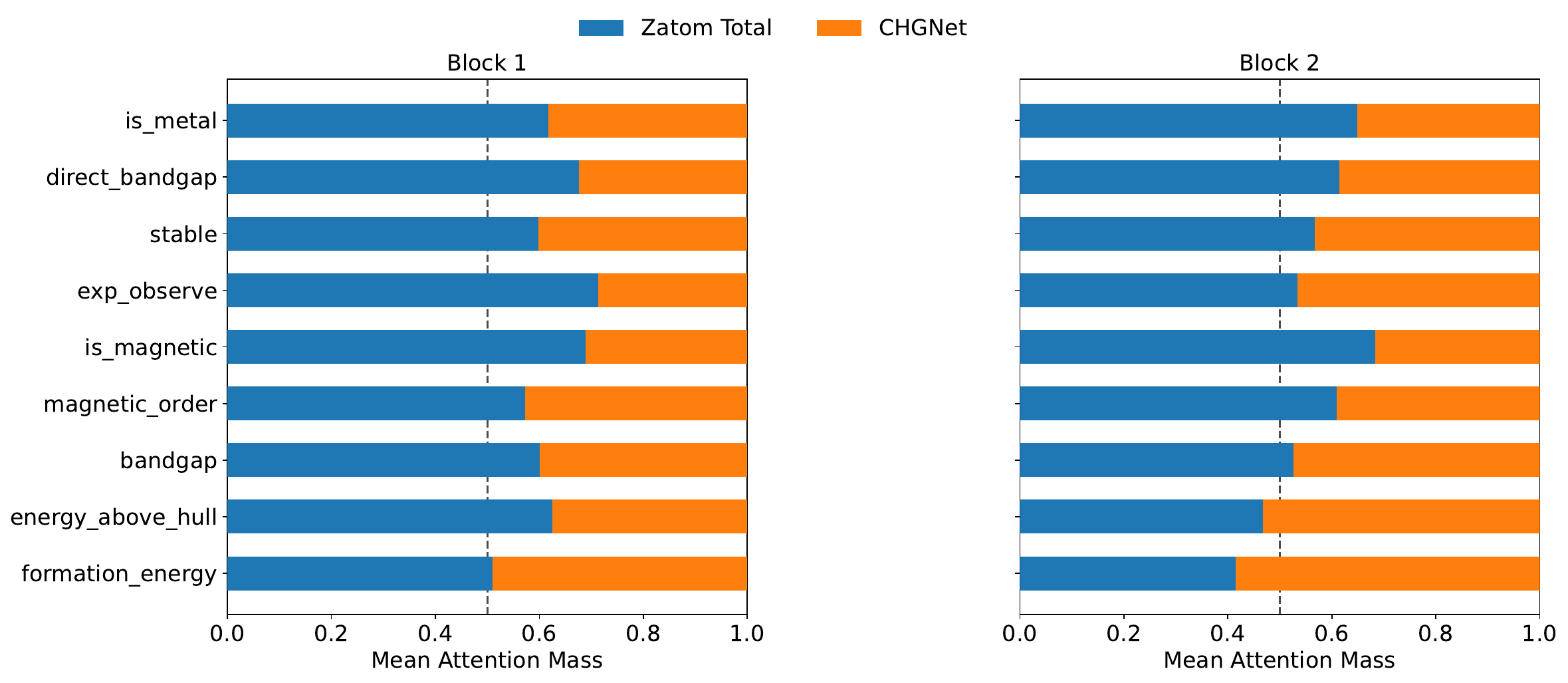}
    \caption{Macro-level encoder contribution (Zatom vs. CHGNet) for the task-conditioned Perceiver model. The stacked bars demonstrate how task-specific latent queries dynamically modulate the balance of cross-attention mass between the aggregated Zatom layers and the CHGNet embedding.}
    \label{fig:sota_stacked_bar}
\end{figure*}

Introducing task conditioning fundamentally alters this behavior, enabling highly dynamic, property-aware feature routing. At the macro-level (\Cref{fig:sota_stacked_bar}), the conditioned TCAF actively modulates its reliance on the two distinct structural encoders depending on the query. For thermodynamic properties such as formation\_energy and energy\_above\_hull, the model leans heavily on the CHGNet embedding, allocating a majority of its attention mass (over 50\% in Block 2) to this source. Conversely, for quantum-mechanical and electronic targets like is\_metal, direct\_bandgap, and is\_magnetic, the Perceiver shifts its focus dramatically toward the Zatom representation, assigning it nearly 70\% of the attention mass in Block 2.

\begin{figure*}[t]
    \centering
    \includegraphics[width=1\textwidth]{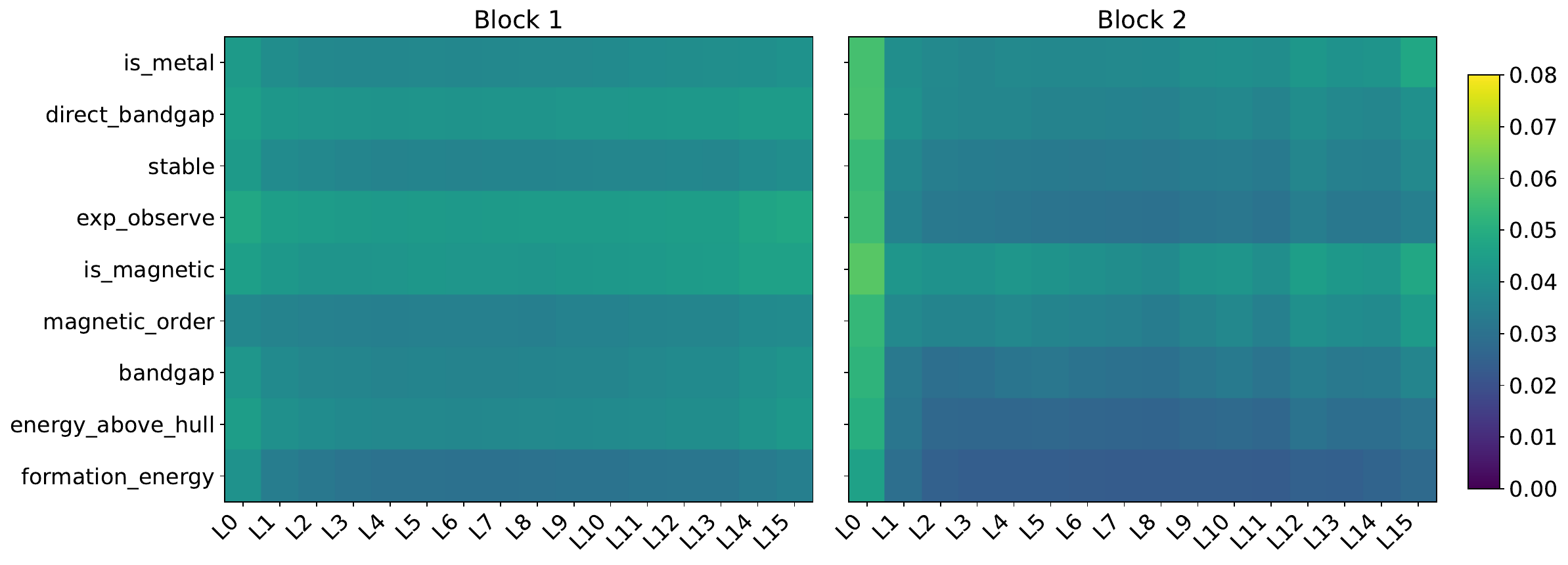}
    \caption{Micro-level cross-attention distribution across the 16 Zatom intermediate layers for the task-conditioned Perceiver model. The heatmap reveals task-aware depth routing, highlighting how the model adaptively extracts structural and chemical features from different depths based on the target property.}
    \label{fig:sota_zatom_heatmap}
\end{figure*}

This adaptive routing is further detailed at the micro-level (\Cref{fig:sota_zatom_heatmap}). Rather than applying a uniform extraction pattern, the task-conditioned model learns unique depth-wise selection strategies. The heatmap illustrates distinct horizontal variations, proving that the queries dynamically navigate the 16 intermediate Zatom layers to pull specific chemical and structural representations at different abstraction depths. This task-aware modulation confirms that TCAF does not merely aggregate data but actively reasons about which encoder, and which specific layer depth, contains the most relevant physical priors for a given target property.

\end{document}